\documentclass[journal]{IEEEtran}

\usepackage{times}
\usepackage{graphicx} 
\usepackage{subfigure} 

\usepackage{algorithm}
\usepackage{algorithmic}

\usepackage{helvet}
\usepackage{courier}

\usepackage{makecell}
\usepackage{graphicx}
\usepackage{subfigure}
\usepackage{color}
\usepackage{amsfonts}
\usepackage{amsmath}
\usepackage{amssymb}

\usepackage{multirow}
\usepackage{booktabs}

\DeclareMathAlphabet\mathbfcal{OMS}{cmsy}{b}{n}

\def\0{{\bf 0}}
\def\1{{\bf 1}}

\usepackage{ntheorem}

\newtheorem*{*thm}{Theorem}

\newtheorem*{*lemma}{Lemma}

\def\ie{\mbox{\textit{i.e.}}}
\def\eg{\mbox{\textit{e.g.}}}

\usepackage{subfigure}
\usepackage{multirow}
\usepackage{url}
\usepackage{cite}
\usepackage{hyperref}

\newcommand{\tabincell}[2]{\begin{tabular}{@{}#1@{}}#2\end{tabular}}

\begin{document}
	
\title{Hierarchical Neural Architecture Search for Single Image Super-Resolution}

	\author{Yong Guo$^*$, Yongsheng Luo$^*$, Zhenhao He$^*$, Jin Huang, Jian Chen$^\dagger$
	\IEEEcompsocitemizethanks{
	\IEEEcompsocthanksitem{Yong Guo, Yongsheng Luo, Zhenhao He, and Jian Chen, are with the School of Software Engineering, South China University of Technology. Jian Chen is also with Guangdong Key Laboratory of BigData Analysis and Processing.
	E-mail: \{guo.yong, seluoyongsheng, sezhenhao.he\}@mail.scut.edu.cn, ellachen@scut.edu.cn}
	\IEEEcompsocthanksitem{Jin Huang is with 
	the School of Computer Science,
	South China Normal University. E-mail: huangjin@m.scnu.edu.cn.
	}
	}
    \thanks{$^*$ Authors contributed equally to this work.}
    \thanks{$^\dagger$ Corresponding author.}
    }

\maketitle

\begin{abstract}
Deep neural networks have exhibited promising performance in image super-resolution (SR). Most SR models follow a hierarchical architecture that contains both the cell-level design of computational blocks and the network-level design of the positions of upsampling blocks. However, designing SR models heavily relies on human expertise and is very labor-intensive. More critically, these SR models often contain a
huge number of parameters and may not meet the requirements of computation resources in real-world applications. To address the above issues, we propose a Hierarchical Neural Architecture Search (HNAS) method to automatically design promising architectures with different requirements of computation cost. To this end, we design a hierarchical SR search space and propose a hierarchical controller for architecture search. Such a hierarchical controller is able to simultaneously find promising cell-level blocks and network-level positions of upsampling layers. Moreover, to design compact architectures with promising performance, we build a joint reward by considering both the performance and computation cost to guide the search process. Extensive experiments on five benchmark datasets demonstrate the superiority of our method over existing methods.
\end{abstract}


\IEEEpeerreviewmaketitle

\section{Introduction}

\IEEEPARstart{I}{mage} super-resolution (SR) is an important computer vision task that aims at designing effective models to reconstruct the high-resolution (HR) images from the low-resolution (LR) images~\cite{freeman2000learning,guo2020closed,yang2019lightweight,yang2019multilevel,huang2019pyramid}.
Most SR models consist of two components, namely several upsampling layers that increase spatial resolution and a set of computational blocks (\eg, residual block) that increase the model capacity. These two kinds of blocks/layers often follow
a two-level architecture, where the network-level architecture determines the positions of the upsampling layers (\eg, SRCNN~\cite{dong2015image} and LapSRN~\cite{lai2017deep}) and the cell-level architecture controls the computation of each block/layer (\eg, RCAB~\cite{zhang2018image}).
In practice, designing deep models is often very labor-intensive
% and relies heavily on human expertise
% ~\cite{guo2019auto,guo2018double,cao2018adversarial,guo2016shallow}. 
% More critically, 
and the hand-crafted architectures are often not optimal in practice.

Regarding this issue, many efforts have been made to automate the model designing process via Neural Architecture Search (NAS)~\cite{pham2018efficient}.
Specifically, NAS methods seek to find the optimal cell architecture~\cite{zoph2018learning,liu2018darts,pham2018efficient,guo2019nat,guo2020breaking} or a whole network architecture~\cite{zoph2016neural,cai2019proxylessnas,tan2019mnasnet,Cai2020Once}. 
% These automatically discovered architectures often outperform the manually designed architectures in both image classification and language modeling tasks~\cite{zoph2016neural,pham2018efficient}.
However, existing NAS methods may suffer from two limitations if we apply them to search for an optimal SR architecture.

First, it is hard to directly search for the optimal two-level SR architecture.
For SR models, both the cell-level blocks and network-level positions of upsampling layers play very important roles.
However, existing NAS methods only focus on one of the architecture levels. Thus, how to simultaneously find the optimal cell-level block and network-level positions of upsampling layers is still unknown.

Second, most methods only focus on improving SR performance but ignore the computational complexity.
{As a result, SR models are often very large and}
become hard to be applied to real-world applications~\cite{zhuang2018discrimination,liu2020discrimination} when the computation resources are limited.
Thus, it is important to design promising architectures with low computation cost.

To address the above issues, we propose a novel Hierarchical Neural Architecture Search (HNAS) method to automatically design SR architectures. Unlike existing methods, HNAS simultaneously searches for the optimal cell-level blocks and the network-level positions of upsampling layers. Moreover, by considering the computation cost to build the joint reward, our method is able to produce promising architectures with low computation cost.

Our contributions are summarized as follows:
\begin{itemize}
    \item We propose a novel Hierarchical Neural Architecture Search (HNAS) method to automatically design cell-level blocks and determine network-level positions of upsampling layers.
    \item We propose a joint reward that considers both the SR performance and the computation cost of SR architectures. By training HNAS with such a reward, we can obtain a series of architectures with different performance and computation cost. 
    \item Extensive experiments on several benchmark datasets demonstrate the superiority of the proposed method.
\end{itemize}

\section{Proposed Method}

\begin{figure*}[!ht]
\centering
\includegraphics[scale=0.35]{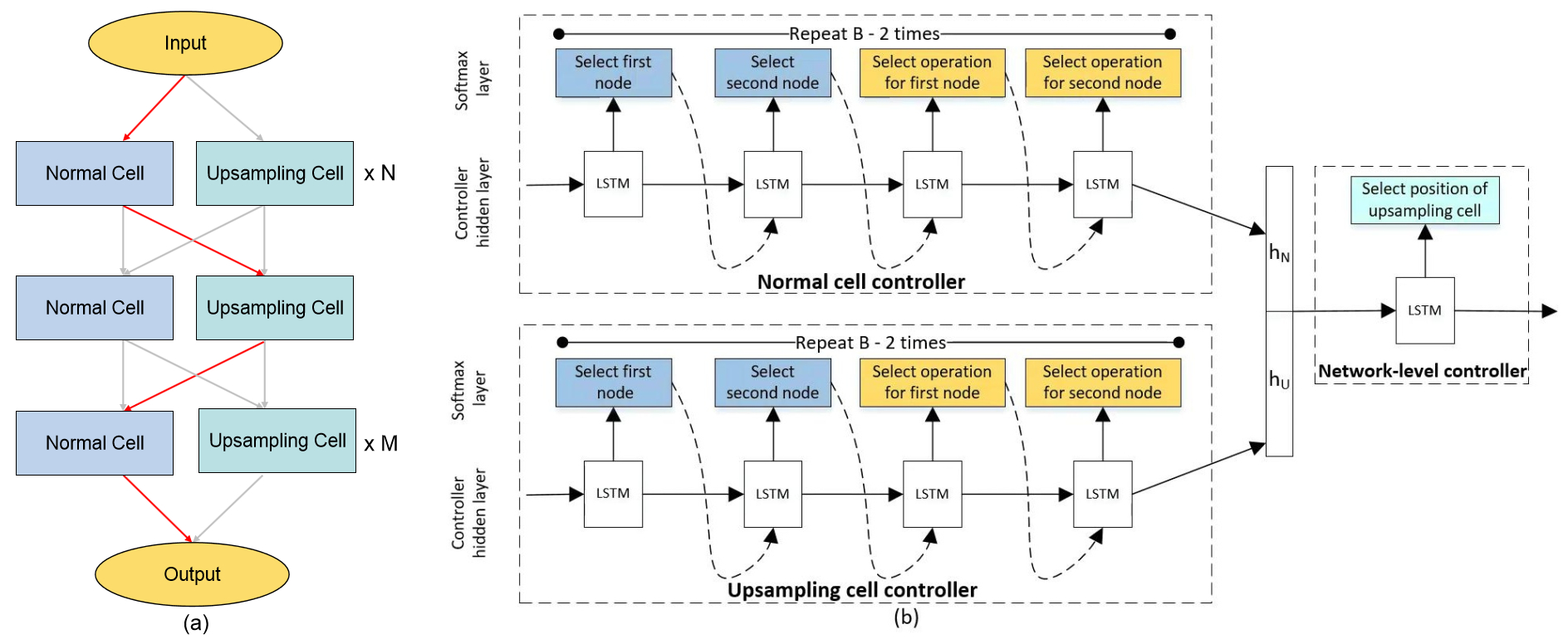}
\vspace{-0.3cm}
\caption{{The overview of the proposed HNAS method. (a) The architecture of the two-branch super network. The red line represents a searched model with the upsampling layer at a specific layer. $N$ and $M$ denote the number of layers before and after the upsampling blocks. (b) The proposed hierarchical controller.}}
\label{overview}
\end{figure*}

In this paper, we propose a Hierarchical Neural Architecture Search (HNAS) method to automatically design promising two-level SR architectures, \ie, with good performance and low computation cost.
To this end, we first define our hierarchical search space that consists of a cell-level search space and a network-level search space.
Then, we propose a hierarchical controller as an agent to search for good architectures. To search for promising SR architectures with low computation cost, we develop a joint reward by considering both the performance and computation cost. We show the overall architecture and the controller model of HNAS in Figure~\ref{overview}.

\begin{figure}
    \centering
    \includegraphics[width=0.75\linewidth]{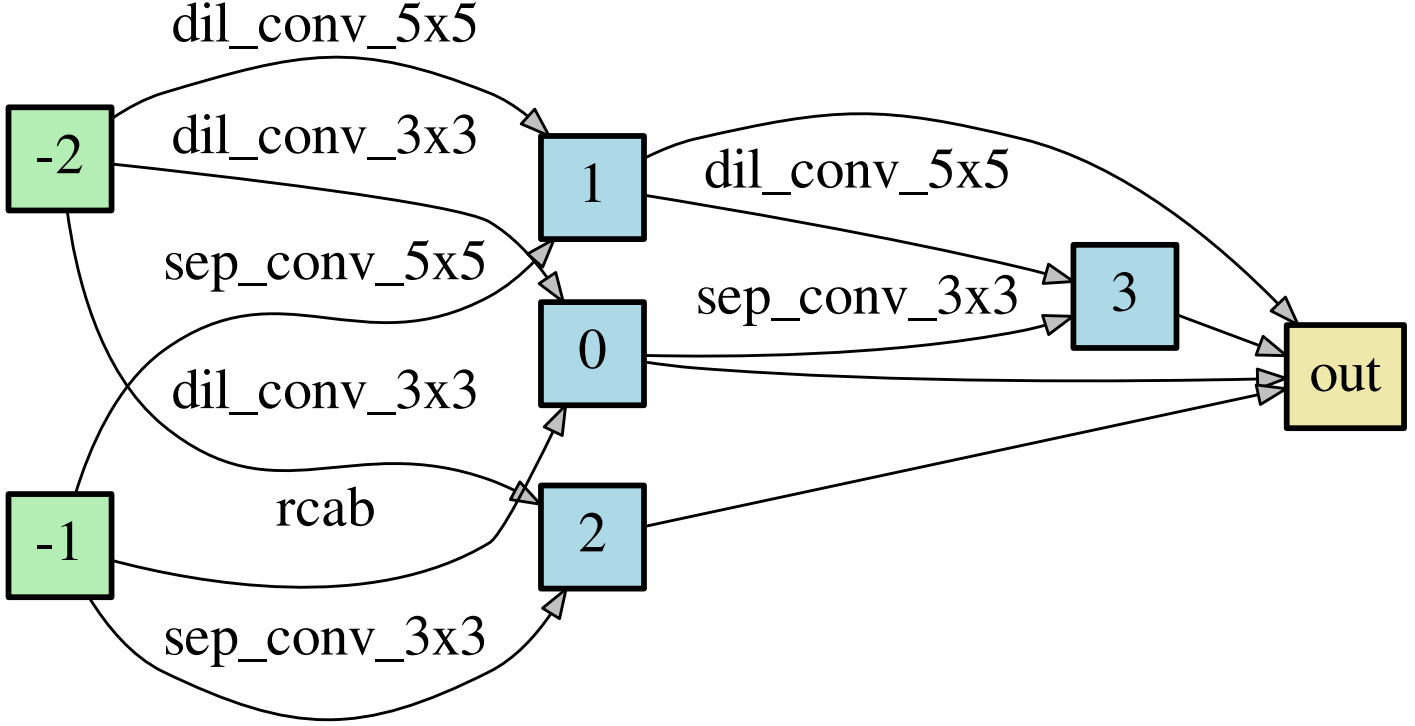}
    \caption{An example of DAG that represents a cell architecture.}
    \label{fig:cell_example}
\end{figure}

\subsection{Hierarchical SR Search Space}
\label{sec41}

In general, SR models often consist of two components, namely several upsampling layers that increase spatial resolution and a series of computational blocks that increase the model capacity.
These two components form a two-level architecture,
where the cell-level identifies the computation of each block and the network-level determines the positions of the upsampling layers.
Based on the hierarchical architecture, we propose a hierarchical SR search space that contains a cell-level search space and a network-level search space.

\textbf{Cell-level search space.}
In the cell-level search space, as shown in Fig.~\ref{fig:cell_example}, we represent a cell as a directed acyclic graph (DAG)~\cite{pham2018efficient,liu2018darts,xu2020pcdarts,chen2019progressive}, where the nodes denote the feature maps in deep networks and the edges denote some computational operations, \eg, convolution. 
In this paper, we define two kinds of cells: (1) the normal cell that controls the model capacity and keeps the spatial resolution of feature maps unchanged,
and (2) the upsampling cell that 
increases the spatial resolution.
To design these cells, we collect the two sets of operations that have been widely used in SR models. We show the candidate operations for both cells in TABLE~\ref{operation}.

For the normal cell, we consider seven candidate operations, including identity mapping, $3 \times 3$ and $5 \times 5$ dilated convolution, $3 \times 3$ and $5 \times 5$ separable convolution, up and down-projection block (UDPB)~\cite{haris2018deep}, and residual channel attention block (RCAB)~\cite{zhang2018image}.
For the upsampling cell, we consider 5 widely used operations to increase spatial resolution.
Specifically, there are 3 interpolation-based upsampling operations, including area interpolation, bilinear interpolation \cite{dong2015image}, nearest-neighbor interpolation \cite{dumoulin2016learned}. Moreover, we also consider 2 trainable convolutional layers, namely the deconvolution layer (also known as transposed convolution)~\cite{zeiler2011adaptive} and the sub-pixel convolution \cite{shi2016real}.

Based on the candidate operations, the goal of HNAS is to select the optimal operation for each edge of DAG and learn the optimal connectivity among nodes (See more details in Section~\ref{sec:controller}).

\begin{table}[t]
	\centering
	\caption{Candidate operations for normal cell and upsampling cell.}
	\begin{tabular}{l|l}
		\toprule  
		Normal Cell/Block & Upsampling Cell/Block \\
		\hline
          $\bullet \text{ identity (skip connection)}$ &
              $\bullet \text{ area interpolation}$ \\
	
		  $\bullet \text{ 3} \times \text{3 dilated convolution}$ &
		  $\bullet \text{ bilinear interpolation}$ \\
		
		  $\bullet \text{ 5} \times \text{5 dilated convolution}$ &
		  $\bullet \text{ nearest-neighbor interpolation}$ \\

          $\bullet \text{ 3} \times \text{3 separable convolution}$ &
          $\bullet \text{ sub-pixel layer}$ \\

          $\bullet \text{ 5} \times \text{5 separable convolution}$ &
          $\bullet \text{ deconvolution layer}$ \\

          $\bullet \text{ up and down-projection block}$~\cite{haris2018deep} & \\

          $\bullet \text{ residual channel attention block}$~\cite{zhang2018image} & \\
		\bottomrule  
	\end{tabular}
	\label{operation}
\end{table}

\textbf{Network-level search space.} 
Note that the position of upsampling block/layer plays an important role in both the performance and computation cost of SR models. Specifically, if we put the upsampling block in a very shallow layer, the feature map would increase too early and hence significantly increase the computational cost of the whole model. By contrast, when we put the upsampling block in a deep layer, there would be little or no layers to process the upsampled features and hence the computation to obtain high-resolution images may be insufficient, leading to suboptimal SR performance. Regarding this issue, we seek to find the optimal position of the upsampling block for different SR models.

{To this end, we design a two-branch super network whose architecture is fixed through the whole search process. As shown in Fig.~\ref{overview}, there are two kinds of cells (\ie, normal cell and upsampling cell) at each layer. Given a specific position of the upsampling cell, we set the selected layer to the upsampling cell and set the others to the normal cells. In this way, the model with a specific position of the upsampling cell becomes a sub-network of the proposed super network.}
Let $N$ and $M$ denote the number of layers before and after the upsampling blocks. 
Thus, there are $L=M+N+1$ blocks in total. 
We will show how to determine the position of the upsampling layer in Section~\ref{sec:controller}.

\subsection{Hierarchical Controller for HNAS}\label{sec:controller}
Based on the hierarchical search space, we seek to search for the optimal cell-level and network-level architectures.
Following~\cite{zoph2016neural,pham2018efficient}, we use a long short-term memory (LSTM)~\cite{hochreiter1997long} as the controller to produce candidate architectures (represented by a sequence of tokens~\cite{zoph2016neural}). 
Regarding the two-level hierarchy of SR models, we propose a hierarchical controller to produce promising architectures.
Specifically, we consider two kinds of controllers, including a cell-level controller that searches for the optimal architectures for both normal block and upsampling block, and a network-level controller that determines the positions of upsampling layers.

\textbf{Cell-level controller.}
We utilize a cell-level controller to find the optimal computational DAG with $B$ nodes (See example in Fig.~\ref{fig:cell_example}).
In a DAG, the input nodes $-2$ and node $-1$ 
denote the outputs of the second nearest and the nearest cell in front of the current block, respectively.
The remaining $B {-} 2$ nodes are intermediate nodes, each of which also takes two previous nodes in this cell as inputs. 
For each intermediate node, the controller makes two kinds of decisions: 1) which previous node should be taken as input and 2) which operation should be applied to each edge. 
All of these decisions can be represented as a sequence of tokens and thus can be predicted using the LSTM controller~\cite{zoph2016neural}.
After repeating $B {-} 2$ times, all of the $B {-} 2$ nodes are concatenated together to obtain the final output of the cell, \ie, the output node.

\textbf{Network-level controller.}
Once we have the normal block and upsampling block, we seek to further determine where we should put the upsampling block to build the SR model.
Given a model with $L$ layers, we predict the position, \ie, an integer ranging from 1 to $L$, where we put the upsampling block. Since such a position relies on the design of both normal and upsampling blocks, we build the network-level controller that takes the embeddings (\ie, hidden states) of two kinds of blocks as inputs to determine the position.
Specifically, let $h_N$ and $h_U$ denote the last hidden states of the controllers for normal block and upsampling block, respectively. 
We concatenate these embeddings as the initial state of the network level controller (See Fig.~\ref{overview}(b)). Since the network-level controller considers the information of the architecture design of both the normal and upsampling blocks, it becomes possible to determine the position of the upsampling block.

\begin{algorithm}[t]
	\caption{\small{Training method for HNAS.}}
    	\begin{algorithmic}[1]\small
    		\REQUIRE The number of iterations $T$, learning rate $\eta$,  shared parameters $w$, controller parameters $\theta$.
            \STATE {Initialize $w$ and $\theta$.}
            \FOR{$i{=}1$ to $T$}
            \STATE // \emph{Update $w$ by minimizing the training loss}
        		\FOR{each iteration on training data}
                    \STATE Sample $\alpha \sim \pi(\alpha;\theta)$;
                    \STATE $w \gets w - \eta \nabla_{w} \mathcal{L}(\alpha,w)$. \\
                \ENDFOR
        		\STATE // \emph{Update $\theta$ by maximizing the reward}
        		\FOR{each iteration on validation data}
                    \STATE Sample $\alpha \sim \pi(\alpha;\theta)$;
            		\STATE $\theta \gets \theta + \eta \mathcal{R}(\alpha) \nabla_\theta\log \pi(\alpha;\theta,\Omega_i)$; \\
            	\ENDFOR
            \ENDFOR
    	\end{algorithmic}
		\label{algorithm1}
\end{algorithm}

\subsection{Training and Inference Methods}

To train HNAS, we first propose the joint reward to guide the architecture search process. Then, we depict the detailed training and inference methods of HNAS.

\textbf{Joint reward.}
Designing promising architectures with low computation cost is important for real-world SR applications.
To this end, we build a joint reward by considering both performance and computation cost to guide the architecture search process.
Given any architecture $\alpha$, let ${\rm PSNR}(\alpha)$ be the PSNR performance of $\alpha$, ${\rm Cost}(\alpha)$ be the computation cost of $\alpha$ in terms of FLOPs (\ie, the number of multiply-add operations). The joint reward can be computed by
\begin{equation}
    \mathcal{R}(\alpha) = \lambda * \text{PSNR}(\alpha) - (1 - \lambda) * \text{Cost}(\alpha),
    \label{eq:reward}
\end{equation}
{where $\lambda$ controls the trade-off between the PSNR performance and the computation cost. Such a trade-off exists when there is a limited budget of computation resources and we can adjust $\lambda$ in the proposed joint reward function to meet different requirements of real-world applications. In general, a larger $\lambda$ makes the controller pay more attention to improving the PSNR performance but regardless of the computation cost. By contrast, a smaller $\lambda$ makes the controller focus more on reducing the computation cost.}

\textbf{Training method for HNAS.} 
With the joint reward, following~\cite{zoph2016neural,pham2018efficient}, we apply the policy gradient \cite{williams1992simple} to train the controller. We show the training method in Algorithm~\ref{algorithm1}. 
To accelerate the training process, we adopt the parameter sharing technique~\cite{pham2018efficient}, \ie, we construct a large computational graph, where each subgraph represents a neural network architecture, hence forcing all architectures to share the parameters. 

Let $\theta$ and $w$ be the parameters of the controller model and the shared parameters. The goal of HNAS is to learn an optimal policy $\pi(\cdot)$ and produce candidate architectures by conduct sampling $\alpha {\sim} \pi(\alpha)$.
To encourage exploration, we introduce an entropy regularization term into the objective.
{In this way, we can train as diverse architectures as possible in the super network and thus alleviate the unfairness issue~\cite{chu2019fairnas} that some sub-networks (or candidate operations) may be over-optimized while the others are under-optimized.}

\begin{table*}[t!]
\centering
\caption{Comparisons with the state-of-the-art methods based on $\times$2 super-resolution task. Results marked with ``$^\dagger$'' were obtained by training the corresponding architectures using our setup. ``-'' denotes the results that are not reported.}
\begin{tabular}{c|c|ccccc}
\toprule
Model & \#FLOPs (G) 
& \tabincell{c}{SET5 \\ PSNR / SSIM}
& \tabincell{c}{SET14 \\ PSNR / SSIM}
& \tabincell{c}{B100 \\ PSNR / SSIM}
& \tabincell{c}{Urban100 \\ PSNR / SSIM}
& \tabincell{c}{Manga109 \\ PSNR / SSIM}
\\

\hline

Bicubic  & - &  33.65 / 0.930 & 30.24 / 0.869 & 29.56 / 0.844 & 26.88 / 0.841 & 30.84 / 0.935 \\
SRCNN \cite{dong2015image}      & 52.7       & 36.66 / 0.954 & 32.42 / 0.906 & 31.36 / 0.887 & 29.50 / 0.894 & 35.72 / 0.968 \\
VDSR \cite{kim2016accurate}     & 612.6        & 37.53 / 0.958 & 33.03 / 0.912 & 31.90 / 0.896 & 30.76 / 0.914 & 37.16 / 0.974 \\
DRCN \cite{kim2016deeply}      & 17,974.3   & 37.63 / 0.958 & 33.04 / 0.911 & 31.85 / 0.894 & 30.75 / 0.913 & 37.57 / 0.973 \\
DRRN \cite{tai2017image}      & 6,796.9      & 37.74 / 0.959 & 33.23 / 0.913 & 32.05 / 0.897 & 31.23 / 0.918 & 37.92 / 0.976 \\
SelNet \cite{choi2017deep}    & 225.7        & 37.89 / 0.959 & \textbf{33.61} / 0.916 & 32.08 / 0.898 & -          & -         \\
CARN \cite{ahn2018fast}      & 222.8         & 37.76 / 0.959 & 33.52 / 0.916 & 32.09 / 0.897 & 31.92 / 0.925 & 38.36 / 0.976 \\
MoreMNAS-A \cite{chu2019multi} & 238.6       & 37.63 / 0.958 & 33.23 / 0.913 & 31.95 / 0.896 & 31.24 / 0.918 & -         \\
FALSR \cite{chu2019fast}     & 74.7          & 37.61 / 0.958 & 33.29 / 0.914 & 31.97 / 0.896 & 31.28 / 0.919 & 37.46 / 0.974 \\
Residual Block~\cite{he2016deep}$^\dagger$ & 47.5 & 36.72 / 0.955 &	32.20 / 0.905	& 31.30 / 0.888&	29.53 / 0.897	& 33.36 / 0.962\\
RCAB~\cite{zhang2018image}$^\dagger$ & 84.9 &37.66 / 0.959&	33.17 / 0.913	& 31.93 / 0.896&	31.19 / 0.918	& 37.80 / 0.974\\
Random$^\dagger$       & 111.7     & 37.83 / 0.959 & 33.31 / 0.915 & 31.98 / 0.897 & 31.42 / 0.920 & 38.31 / 0.976 \\
\hline
HNAS-A ($\lambda=0.2$)      & \textbf{30.6}           & 37.84 / 0.959 & 33.39 / 0.916 & 32.06 / 0.898 & 31.50 / 0.922 & 38.15 / 0.976 \\
HNAS-B ($\lambda=0.6$)      & 48.2          & 37.92 / 0.960 & 33.46 / 0.917 & 32.08 / 0.898 & 31.66 / 0.924 & 38.46 / 0.977 \\
HNAS-C ($\lambda=0.9$)      & 83.6          & \textbf{38.11} / \textbf{0.964} & 33.60 / \textbf{0.920} & \textbf{32.17} / \textbf{0.902} & \textbf{31.93} / \textbf{0.928} & \textbf{38.71} / \textbf{0.985}\\
\bottomrule
\end{tabular}
\label{exp1}
\end{table*}

\textbf{Inferring Architectures.}
Based on the learned policy $\pi(\cdot)$,
we conduct sampling to obtain promising architectures. Specifically, we first sample several candidate architectures and then select the architecture with the highest validation performance. Finally, we build SR models using the searched architectures (including both the cell-level blocks and network-level position of upsampling blocks) and train them from scratch.

\section{Experiments}
\label{exp_set}

In the experiments, we use the DIV2K dataset \cite{timofte2017ntire} to train all the models and conduct comparisons on five benchmark datasets, including Set5~\cite{bevilacqua2012low}, Set14~\cite{zeyde2010single}, BSD100~\cite{martin2001database}, Urban100~\cite{huang2015single}, and Manga109~\cite{fujimoto2016manga109}. 
We compare different models in terms of PSNR, SSIM, and FLOPs.
Please see more training details in supplementary.
We have made the code of HNAS available at \href{https://github.com/guoyongcs/HNAS-SR}{https://github.com/guoyongcs/HNAS-SR}.

\begin{figure}[t]
\centering
\includegraphics[width = 1\columnwidth]{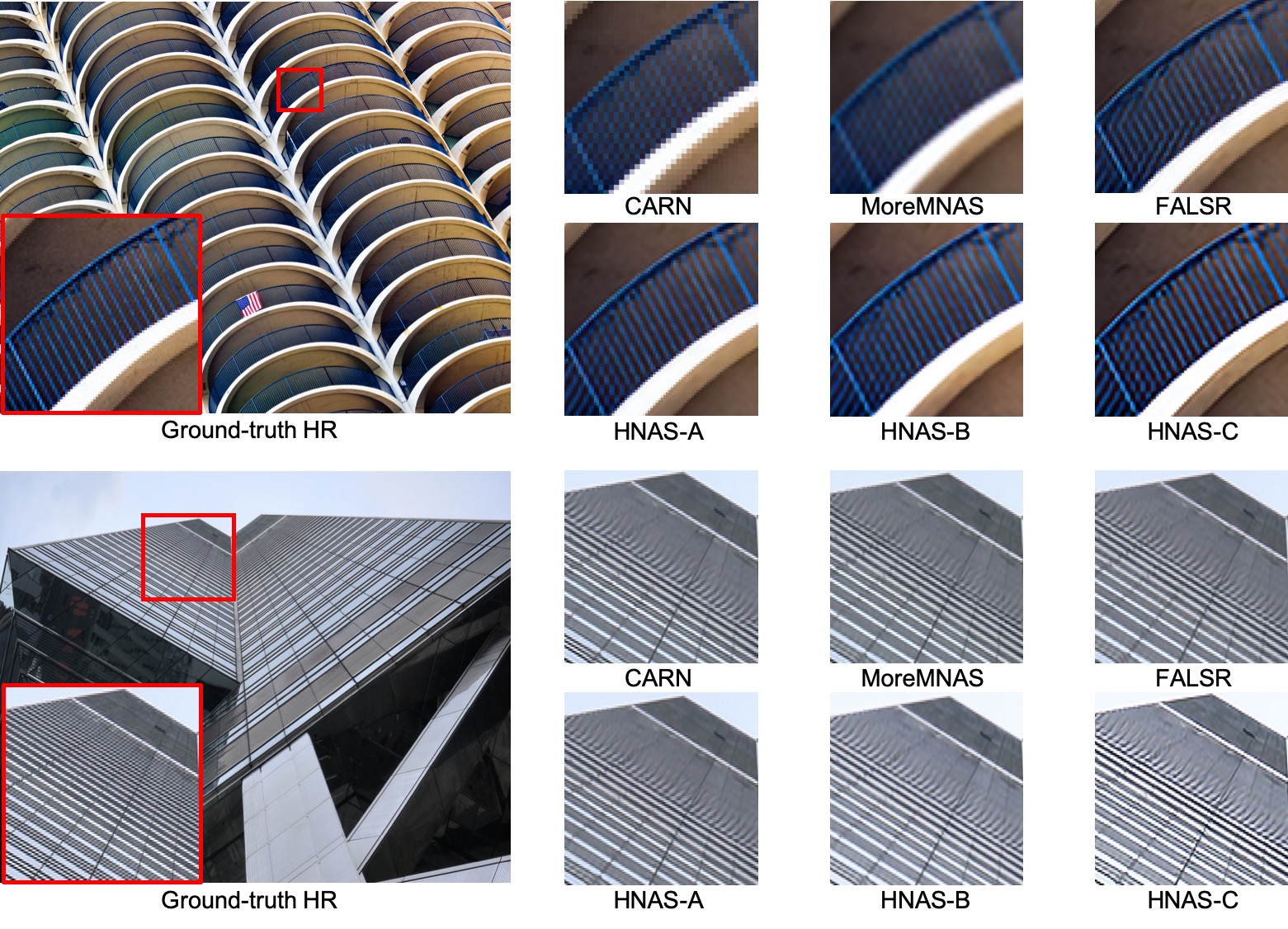}
\caption{Visual comparisons of different methods for $2 \times$ SR.}
\label{exp2}
\end{figure}

\subsection{Quantitative Results}
In this experiment, we {consider three settings (\ie, $\lambda=\{0.2, 0.6, 0.9\}$)} 
{and use HNAS-A/B/C to represent the searched architectures under these settings, respectively.}
We show the detailed architectures in supplementary.
Table~\ref{exp1} shows the quantitative comparisons for $2\times$ SR. Note that all FLOPs are measured based on a $3 \times 480 \times 480$ input LR image. Compared with the hand-crafted models, our models tend to yield higher PSNR and SSIM and lower fewer FLOPs. 
Specifically, HNAS-A yields the lowest FLOPs but still outperforms a large number of baseline methods. 
Moreover, when we gradually increase $\lambda$, HNAS-B and HNAS-C take higher computation cost and yield better performance.
These results demonstrate that HNAS can produce architectures with promising performance and low computation cost.

\subsection{Visual Results}

To further show the effectiveness of the proposed method, we also conduct visual comparisons between HNAS and several state-of-the-arts.
We show the results in Fig.~\ref{exp2}.
From Fig.~\ref{exp2}, the considered baseline methods often produce very blurring images with salient artifacts. By contrast, the searched models by HNAS are able to produce sharper images than other methods.
These results demonstrate the effectiveness of the proposed method.

\section{Conclusion}
In this paper, we have proposed a novel Hierarchical Neural Architecture Search (HNAS) method to automatically search for the optimal architectures for image super-resolution (SR) models. Since most SR models follow the two-level architecture design, we define a hierarchical SR search space and develop a hierarchical controller to produce candidate architectures. Moreover, we build a joint reward by considering both SR performance and computation cost to guide the search process of HNAS. With such a joint reward, HNAS is able to design promising architectures with low computation cost.
Extensive results on five benchmark datasets demonstrate the effectiveness of the proposed method.

\section*{Acknowledgement}
This work was partially supported by the Guangdong Basic and Applied Basic Research Foundation (No. 2019B1515130001), the Guangdong Special Branch Plans Young Talent with Scientific and Technological Innovation (No. 2016TQ03X445), the Guangzhou Science and Technology Planning Project (No. 201904010197) and Natural Science Foundation of Guangdong Province (No. 2016A030313437).

\bibliographystyle{IEEEtran}
\IEEEtriggeratref{22}
\bibliography{ref}

\onecolumn

\begin{LARGE}
	~~~\vspace{1pt}
	\begin{center}
		\bf Supplementary Materials for ``Hierarchical Neural Architecture Search for Single Image Super-Resolution''
	\end{center}
\end{LARGE}
\vspace{2pt}

We organize our supplementary materials as follows. First, we show the concrete structure about our derived models, including HNAS-A, HNAS-B and HNAS-C, and  then do some brief analysis in Section A. Second, we will depict the implementation details about our HNAS models, including the datasets and training details in Section B.

\subsection{Analysis of Model Structures}
\begin{figure}[h]
	\centering
	\subfigure[normal cell of HNAS-A]{\includegraphics[width=0.25\textwidth]{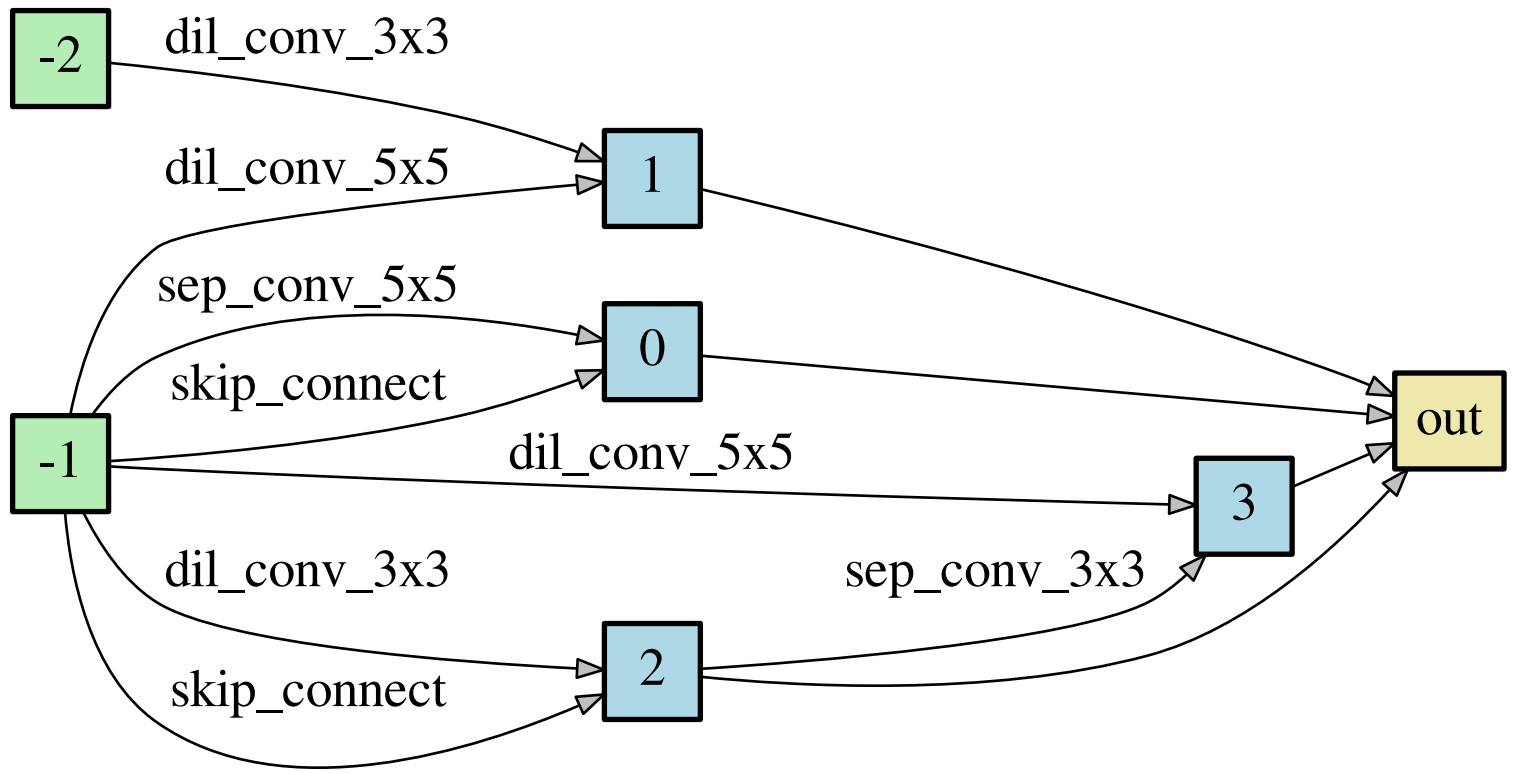}} \quad
	\subfigure[normal cell of HNAS-B]{\includegraphics[width=0.25\textwidth]{SRNAS-B_1.png}} \quad 
	\subfigure[normal cell of HNAS-C]{\includegraphics[width=0.40\textwidth]{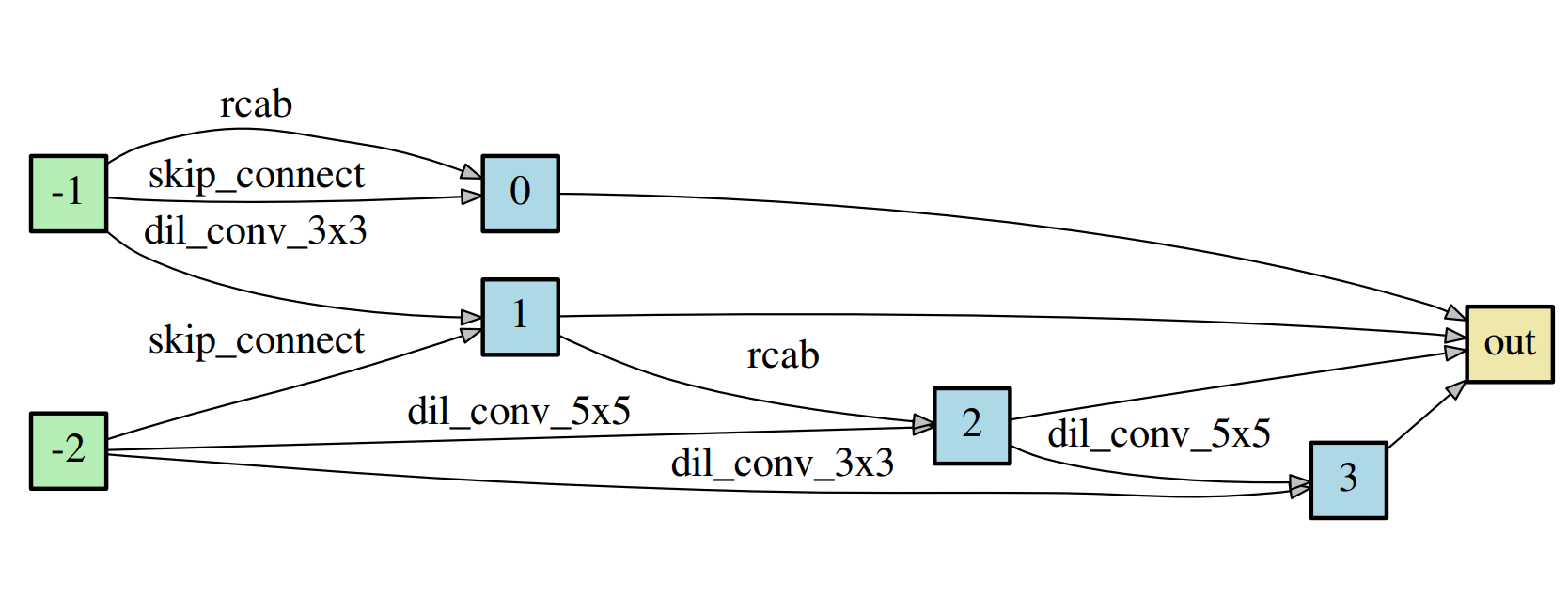}} \quad
	
	\subfigure[upsampling cell of HNAS-A]{\includegraphics[width=0.30\textwidth]{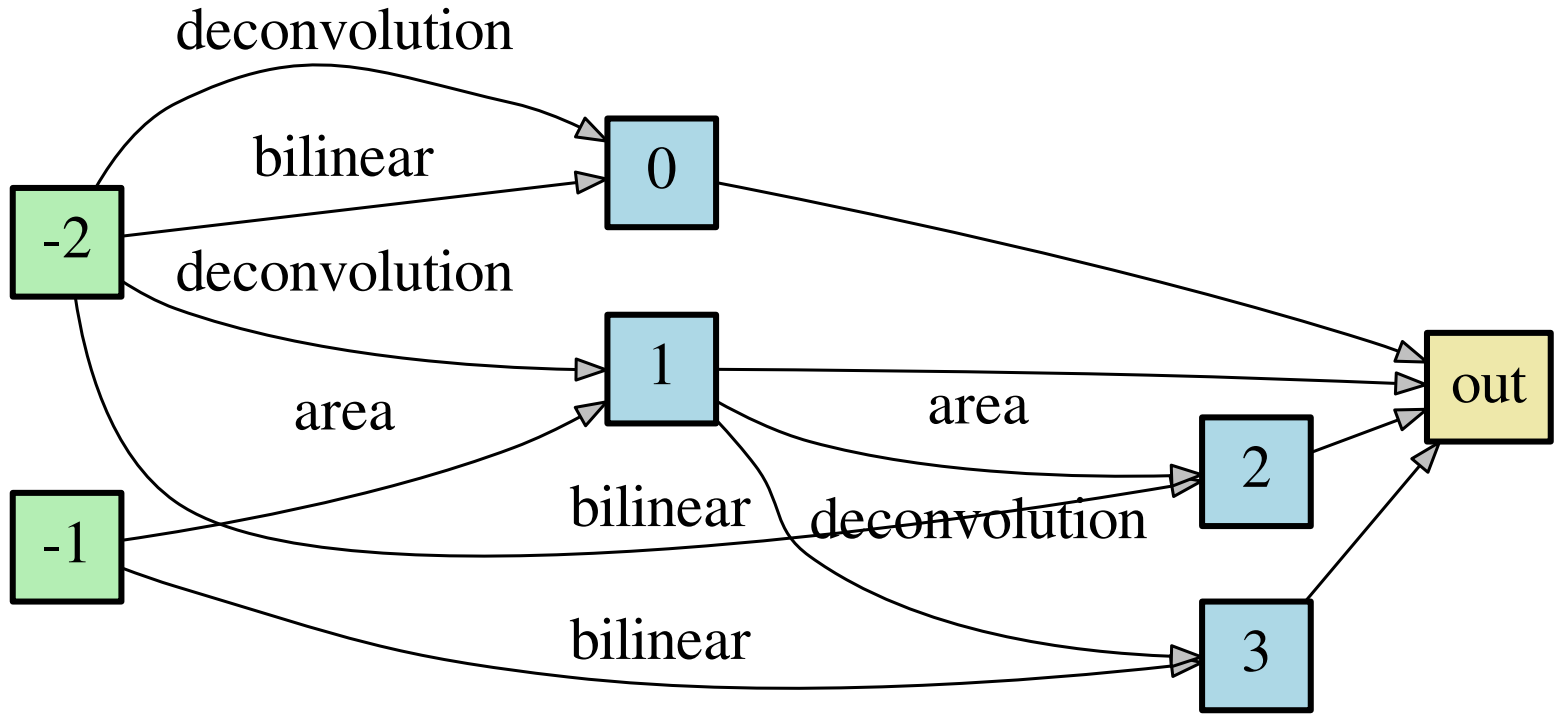}} \quad
	\subfigure[upsampling cell of HNAS-B]{\includegraphics[width=0.30\textwidth]{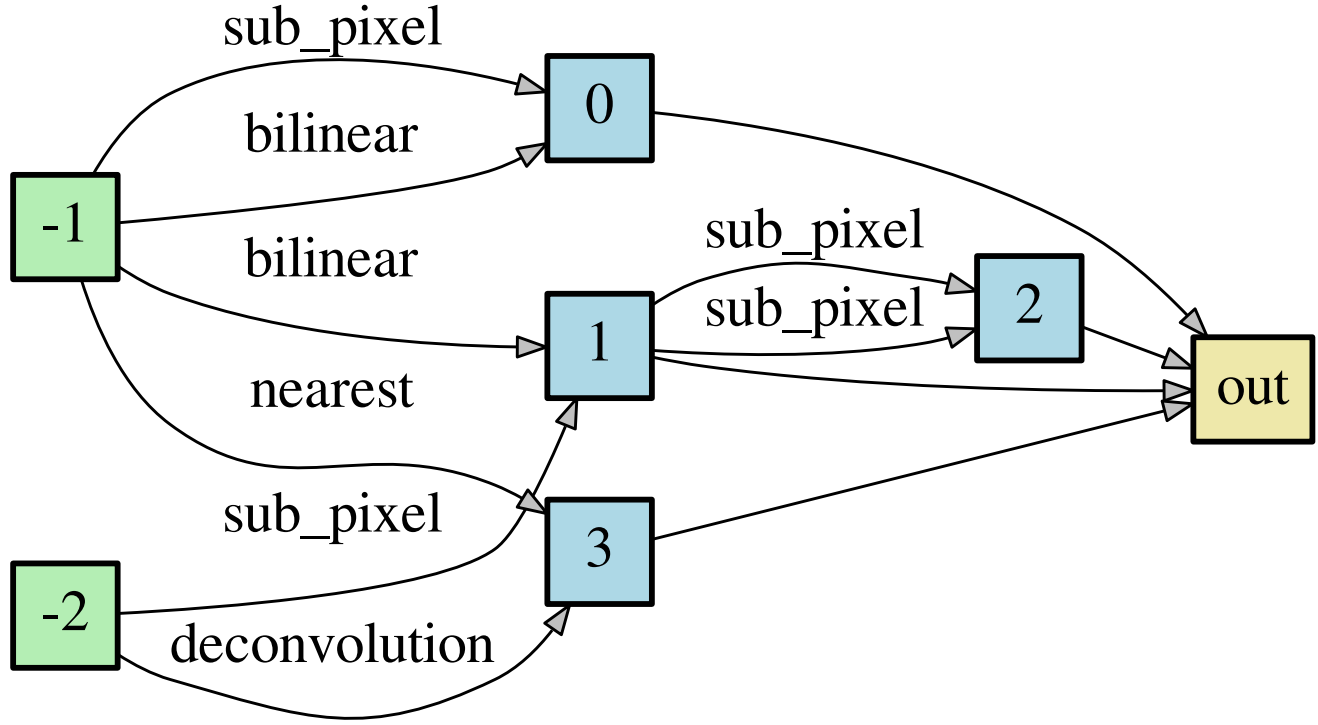}} \quad  
	\subfigure[upsampling cell of HNAS-C]{\includegraphics[width=0.30\textwidth]{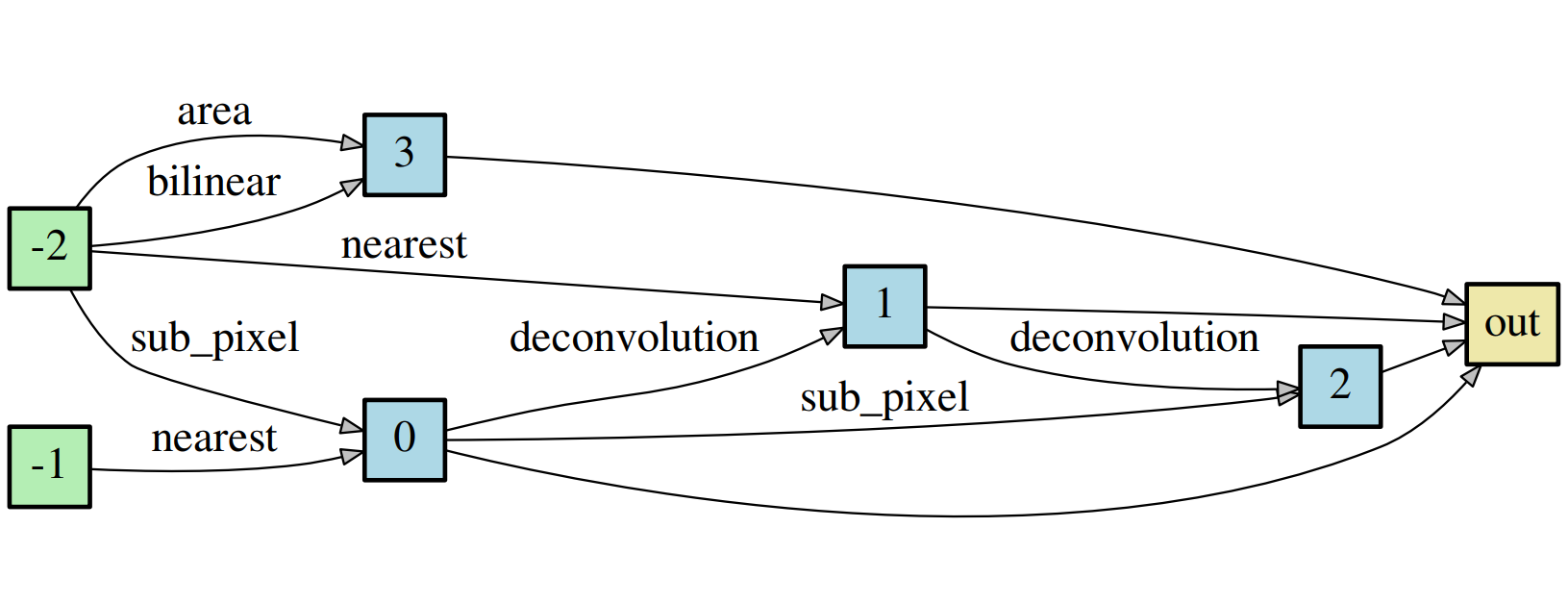}} 
	\caption{The structural details of HNAS-A/B/C. The positions of the upsampling cell for these architectures are 10/12/12 (out of $L=12$ layers), respectively.}
	\label{model_detail}
\end{figure}

\textbf{The Derived Models.}
The graph representations of HNAS-A, HNAS-B and HNAS-C are shown in Fig. \ref{model_detail}. 

As shown in Fig. \ref{model_detail}, we can observe that the structures of our derived models, HNAS-A, HNAS-B and HNAS-C, are quite different from each other.
We provide enough operations for cell nodes to choose, including seven candidate operations and five upsampling operations in the search space
of normal cell and upsampling cell respectively, as mentioned in TABLE~\ref{operation}. Besides, the computational DAG consists of B nodes, and the intermediate B-2 nodes can randomly 
take two previous nodes in this cell as inputs, which provides a series of models that have different structures and performance for model selection.

\subsection{Implementation Details}
When training HNAS model, we use different setting about datasets and hyper-parameters. Thus, we will present  our implementation details on both the datasets and the training details in this section.

\textbf{Datasets.} Following \cite{kim2016accurate}, we train all SR networks using 800 training images from DIV2K dataset \cite{timofte2017ntire}. Specially, at search stage, we split DIV2K training set into 40\% and 60\% slices to train the model parameters $w$  and the transformer parameters $\theta$, respectively.  Set5 \cite{bevilacqua2012low} dataset is used as the validation set. After searching stage, we can infer some novel different cell architectures.  Then, we re-train these cell architectures from scratch with all the 800 images from DIV2K dataset.

\textbf{Training details.} We employ one-layer Long Short-Term Memory (LSTM) \cite{hochreiter1997long} network with 100 as our RNN model. The learning rate  follows a cosine annealing schedule with $\eta_{max}=0.001$ and $\eta_{min}=0.00001$. The batch-size is set to 16. The ADAM optimizer is used to train our model. We train the three LSTM controllers using the same settings, except for using $\eta_{max}=0.0003$ and $\eta_{min}=0.00015$.
We add the controller's sample entropy to the reward, weighted by 1.  Each architecture search is run for 400 epochs. The total number of layer is set to 12 and the channels of each operations to 8. Note that after concatenating the output of cell nodes, the channels of a model is equal to 32. At the inferring stage, we retrain the selected model using the same settings as the searching stage, except for setting 64 channels of each operations.

\end{document}